\documentclass[9pt]{article} 
\usepackage{spconf,amsmath,graphicx}

\usepackage{amsmath,amssymb,amscd,amsthm,amsfonts}
\usepackage{booktabs} 
\usepackage{nicefrac}
\usepackage{mathtools}
\usepackage{bm}
\usepackage{lipsum}
\usepackage{hyperref}
\hypersetup{colorlinks=true, urlcolor=blue}
\usepackage{subfig}
\makeatletter
\newcommand{\printfnsymbol}[1]{%
  \textsuperscript{\@fnsymbol{#1}}%
}
\makeatother

\newcommand{\x}{\mathbf{x}}
\newcommand{\y}{\mathbf{y}}
\newcommand{\z}{\mathbf{z}}
\newcommand{\si}{\mathbf{s}}

\newcommand{\N}{\mathcal{N}}

\title{Gluformer: Transformer-Based Personalized Glucose Forecasting with Uncertainty Quantification}
\name{
Renat Sergazinov$^{*}$, Mohammadreza Armandpour$^{*}$, Irina Gaynanova
}
\address{
Department of Statistics, Texas A\&M University
}
\begin{document}
\maketitle
\def\thefootnote{*}\footnotetext{These authors contributed equally to this work.}\def\thefootnote{\arabic{footnote}}
\begin{abstract}
Deep learning models achieve state-of-the art results in predicting blood glucose trajectories, with a wide range of architectures being proposed. However, the adaptation of such models in clinical practice is slow, largely due to the lack of uncertainty quantification of provided predictions. In this work, we propose to model the future glucose trajectory conditioned on the past as an infinite mixture of basis distributions (i.e., Gaussian, Laplace, etc.). This change allows us to learn the uncertainty and predict more accurately in the cases when the trajectory has a heterogeneous or multi-modal distribution. To estimate the parameters of the predictive distribution, we utilize the Transformer architecture. We empirically demonstrate the superiority of our method over existing state-of-the-art techniques both in terms of accuracy and uncertainty on the synthetic and benchmark glucose data sets.
\end{abstract}
\section{Introduction}
\label{sec:intro}

Prediction of blood glucose values in patients with diabetes is an active area of research~\cite{oviedo2017review}. From a clinical standpoint, accurate forecasting of glucose levels can help patients take proactive actions and prevent severe complications such as hypoglycemia or diabetic coma \cite{sun2021idf, broll2021interpreting}. From the methodological perspective, the highly non-linear and non-stationary nature of the glucose profiles makes accurate predictions difficult. To the best of our knowledge, most current methods produce point-wise predictions without quantifying their uncertainty, which limits clinical applicability. A  predicted glucose increase may prompt the patient to take extra insulin, which may lead to dangerous hypoglycemia if the prediction is wrong. Complementing predictions with uncertainty quantification will help to minimize these risks.
 
 

Since existing approaches primarily focus on maximizing the accuracy of the predictions (rather than quantifying uncertainty), they focus on minimizing the mean squared error (MSE) between the predictions and the observed data \cite{eren2010hypoglycemia, boiroux2012overnight, sudharsan2014hypoglycemia, otoom2015real, novara2015nonlinear,   georga2015evaluation, hidalgo2017data, foxDeepMultiOutputForecasting2018, armandpourDeepPersonalizedGlucose2021}. MSE loss implicitly corresponds to modeling future trajectory as a function of the past plus a homogeneous Gaussian noise. If the assumed model is correct, this approach leads to the best predictions, and confidence intervals can be constructed by inferring the corresponding noise variance. However, a homogeneous Gaussian noise assumption is typically violated in practice, making the corresponding predictions and confidence intervals unreliable. 

Here we propose a new framework for modeling future glucose trajectories using an infinite mixture model (IMM). Such formulation allows us to capture a variety of non-Gaussian output distributions, e.g.,  multi-modal, skewed. We estimate the parameters of the IMM using the Transformer architecture. While IMM has an infinite number of parameters, we introduce an approximation step to fit the distribution implicitly. Specifically, we inject randomness into the computation of the network outputs and treat each output as a realization from one of the mixture components. 


\begin{figure}[t]
\centering
\subfloat{
    \includegraphics[clip, width=0.45\textwidth]{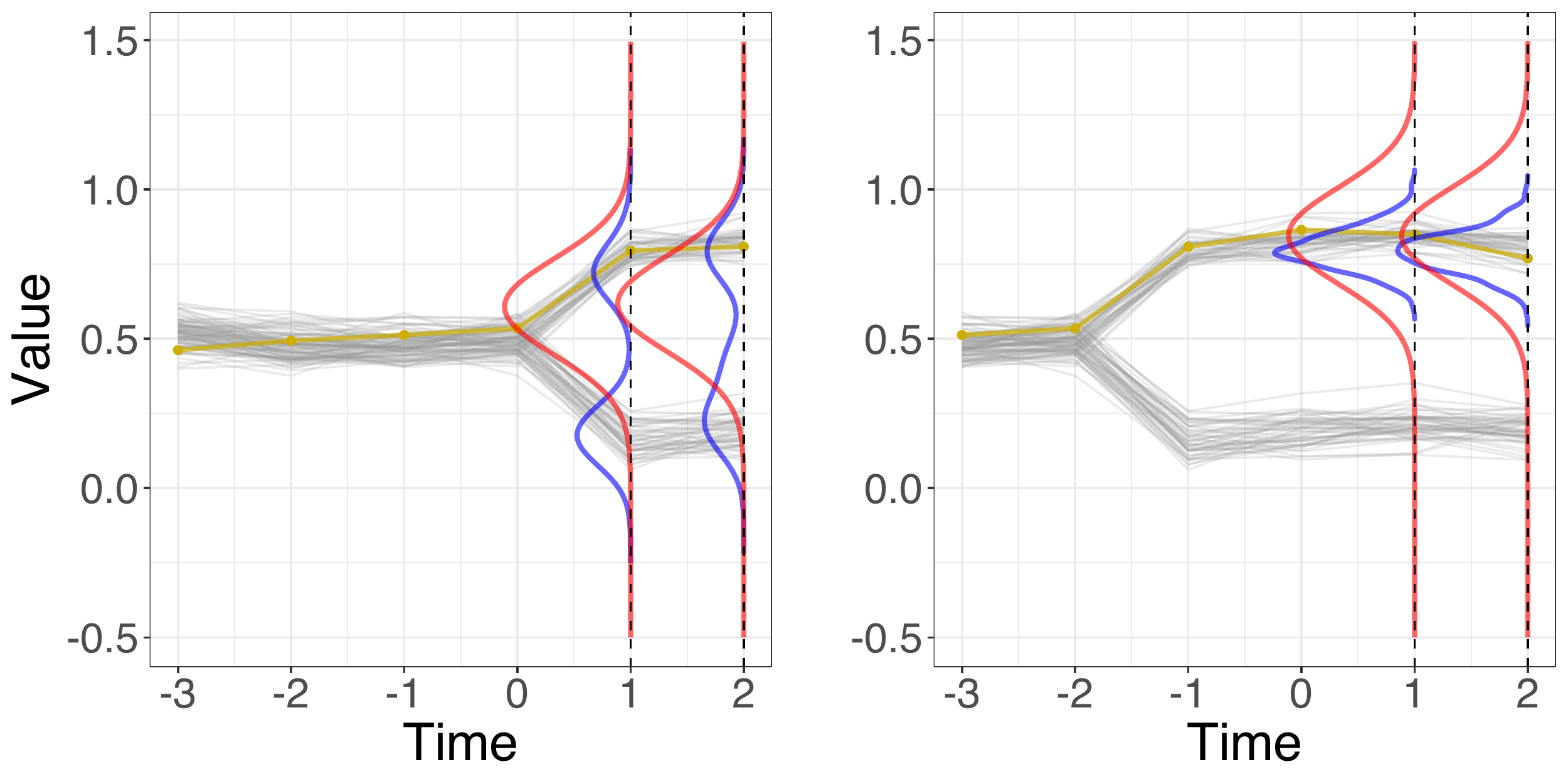}%
}\\
\subfloat{
  \includegraphics[clip, width=0.45\textwidth]
  {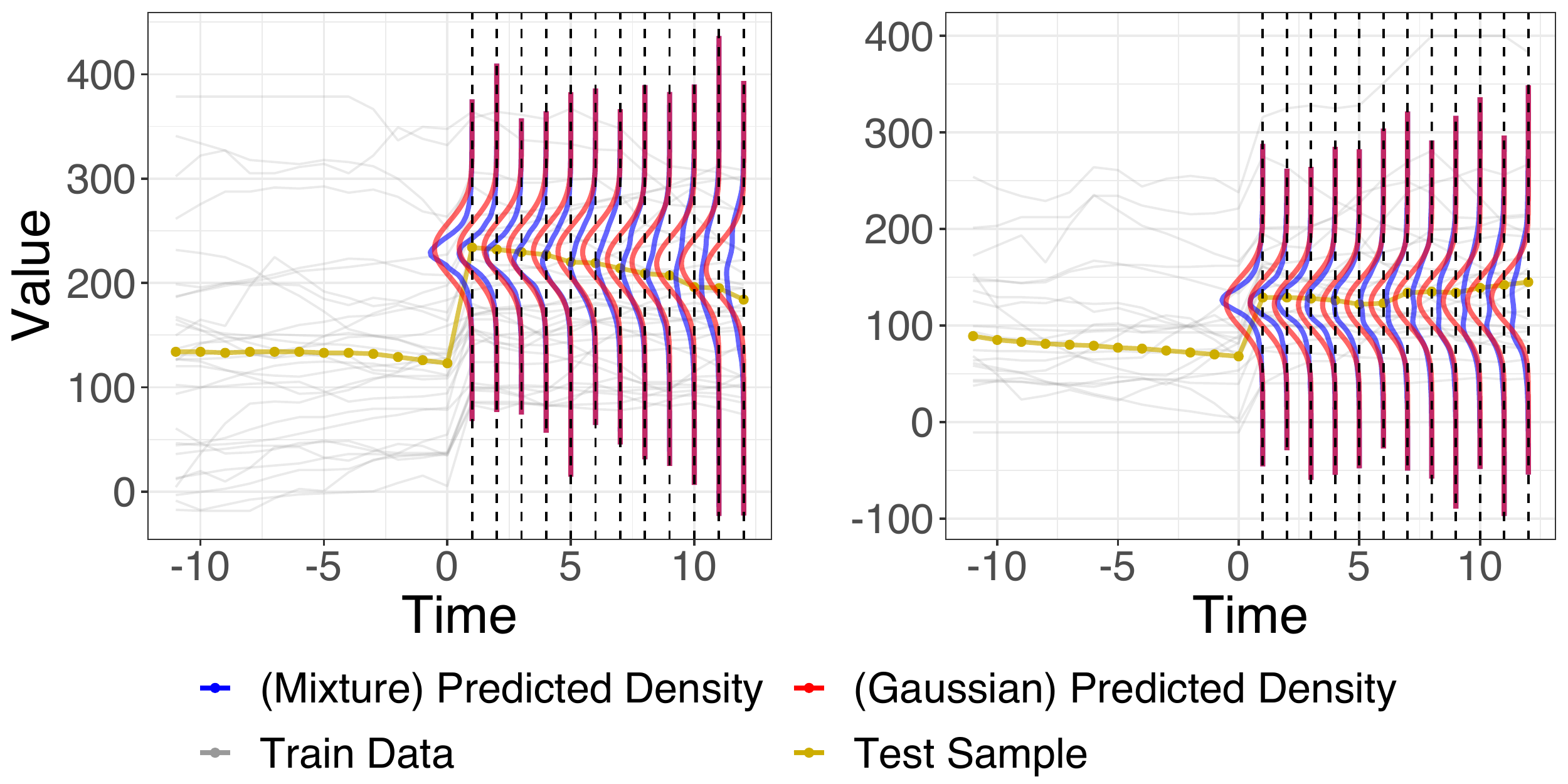}%
}

\caption{Predicted densities on a test sample for synthetic data set \textbf{(top)} and glucose data \textbf{(bottom)}.}
\label{fig:prediction}
\end{figure}


To validate the capabilities of the proposed model, we consider two data sets: 1) synthetic data set 2) publicly available continuous glucose monitor (CGM) data set \cite{foxDeepMultiOutputForecasting2018}. We achieve state-of-the-art results for both. Figure~\ref{fig:prediction} shows a sample of predicted curves together with the visualization of predictive distribution, demonstrating the superiority of our method in terms of the robustness of uncertainty estimates to deviations from the Gaussian assumption.

\section{Related Work}
\label{sec:prev-approaches}

\textbf{Traditional machine learning.} ARIMA is a well-known model for time series data and is commonly used with CGM data \cite{eren2010hypoglycemia, boiroux2012overnight, otoom2015real, novara2015nonlinear}. ARIMA is a non-linear auto-regressive approach that extends the standard AR model to handle non-stationarity by integrating the previous trends. Another popular model for glucose predictions is random forest \cite{sudharsan2014hypoglycemia, georga2015evaluation, hidalgo2017data}. The model splits the data into random subsamples, fitting a single prediction tree for each. The average of the ensemble of individual trees is used for prediction. The model is typically applied recursively by producing a single prediction at a time, but multi-output extensions are also available.


\noindent
\textbf{Deep learning.} 
As time-series resemble data from the natural language processing field, there has been an interest in applying deep learning for multi-output time-series forecasting. Some existing state-of-the-art approaches for glucose prediction are PolySeqMO \cite{foxDeepMultiOutputForecasting2018} and the  RNN-based model \cite{armandpourDeepPersonalizedGlucose2021}. PolySeqMo models the predicted curve with a polynomial expansion and uses an RNN to learn the coefficients. 
The RNN approach of \cite{armandpourDeepPersonalizedGlucose2021} improves PolySeqMO by moving away from the polynomial expansion and building upon the approach taken in \cite{salinas2020deepar}, providing personalized predictions, proposing attention for longer inputs, and introducing robust training scheme for the model. Recently, the Temporal Fusion Transformer (TFT) has shown promising results on a variety of time series data sets \cite{lim2019temporal}. The model is formulated using quantile regression and allows for uncertainty quantification.

\noindent
\textbf{Uncertainty estimation.} The infinite mixture model (IMM) has been previously studied in \cite{rasmussen1999infinite, porteous2012gibbs, yerebakan2014infinite}. These approaches rely on MCMC sampling from the posterior distribution, with latent variables explicitly being sampled from specified prior. 
In contrast, we estimate the IMM parameters implicitly by injecting latent variables (noise) into the network. This change allows us to translate the sampling problem into the maximum-likelihood optimization, where the only additional cost we pay is several stochastic forward passes through the network. Since full MCMC sampling is costly to implement and diagnose for high-dimensional data sets, our approach yields a significant computational advantage.

Stochastic neural networks have been investigate by \cite{gal16} and \cite{depeweg2017learning}. \cite{gal16} explore a connection between a stochastic neural network with a dropout and $L^2$ regularization and the Gaussian process. Specifically, they show that the dropout imposes a distribution on the weights of the neural network, which can then be optimized as a variational approximation to the Gaussian kernel parameters. Their derivations can be seen as a special case of our model if we take the basis distribution to be Gaussian, impose $L^2$ constraint on the weights, and implement the latent variables as a dropout. 



\section{Proposed model}
\label{sec:uncertainty}

\subsection{Notation and background}
\label{sec:formulation}
The blood glucose prediction can be viewed as a time series forecasting problem. Let $\x_i = (x_{ij})_{j=1}^{t}$  be a sequence of glucose values of length $t$ with $i = \{1, \dots, n\}$ being the sample index. Denote the consecutive future glucose values as $\y_i = (y_{ij})_{i=1}^{T}$, so that $(x_{i1}, \dots, x_{it}, y_{i1}, \dots, y_{iT})$ forms an equally-spaced time sequence of CGM readings. Each such sequence corresponds to a unique patient in the data set. The goal is to estimate $\y$ given history $\x$.

To achieve this goal, previous works (Section~\ref{sec:prev-approaches}) focus on building a predictive model $f_{\bm{\theta}}:\mathbb{R}^t\to \mathbb{R}^T$ with parameters $\bm{\theta}$. These models usually rely on the parametric assumption $\y \sim \N(f_{\bm{\theta}}(\x), \sigma^2 I)$, which is equivalent to the additive Gaussian assumption on the residuals. While this assumption is sometimes explicit \cite{salinas2020deepar}, it is often implicit through the choice of the loss function. When the true predictive distribution, $p(\y | \x)$, is far from the Gaussian, such models are clearly misspecified. Subsequently, both predictions and uncertainty estimates of such models become unreliable.

\subsection{The Infinite Mixture Model}
\label{subsec:imm}
We first consider univariate prediction and denote the scalar response variable as $y$. We introduce latent (hidden) variables $\z \sim q$, where $q$ is some distribution that allows fast sampling. Provided we have sampled $\z$ from $q$, we assume that the base distribution assumption holds conditionally on $\z$:
\begin{align*}
    y \lvert \x, \z \sim p(y; \si(\x, \z)),
\end{align*}
where $\si$ is a vector of the sufficient statistics for a distribution $p$, e.g., mean and variance for Gaussian. Marginalizing over $\z$, we get the full predictive distribution expressed as an infinite mixture model (IMM):
\begin{align*}
    p(y \lvert \x) = \int p(y; \si(\x, \z)) q(\z) d\z
\end{align*}
We propose to estimate the sufficient statistics $\si$ with a neural network model $\hat \si \coloneqq f_{\bm{\theta}}(\x | \z)$. To approximate the intractable predictive distribution, we use Monte Carlo integration. That is, we sample $k$ latent $z_j$ independently from $q$, and average over $k$ stochastic passes through~the~network:
$$
\z_j \sim q, \quad 
\hat \si_j = f_{\bm{\theta}}(\x  \lvert \z_j), 
\quad p (y\lvert \x) \approx \frac{1}{k}\sum_{j=1}^k p(y; \hat \si_j).
$$
The latent distribution, $q$, can be implemented by adding a random noise or concatenating a latent noise vector to the input vector, $\x$. Alternatively, the latent distribution may be embedded in the network structure through the Bernoulli dropout layers. The choice of the base distribution, $p$, ultimately depends on the specifics of the problem. While the Gaussian base distribution is a good starting point due to its theoretical properties \cite{rasmussen1999infinite}, Laplace base distribution may be preferable when modeling highly asymmetric distributions \cite{brando2019modelling}. 

To generalize from the univariate case to the multi-output forecasting problem, we model the multivariate predictive distribution $p(\y | \x)$ by specifying univariate IMM models as above for each output and combining them using an appropriate copula dependence function (e.g., independence copula, Gaussian copula etc.). Since, by Sklar's theorem, any multivariate distribution can be decomposed into its marginals and a dependence function, our specification remains expressive \cite{sklar1959fonctions}. The copula function parameters can then also be estimated using the neural network. In practice, we found that the independence copula, corresponding to the independence across time assumption, was sufficient.

\subsection{Measuring the Quality of Predictive Distribution Fit}
\label{subsec:quality}
\textbf{Likelihood.} To assess the quality of estimated predictive distributions, we compute log-likelihoods for each method on the test data, where
a higher value indicates a better fit. 

For existing models (Section~\ref{sec:prev-approaches}), the additive Gaussian assumption on the noise leads to the predictive distribution $\y | \x \sim \N(f_{\bm{\theta}}(\x), \sigma^2 I)$. Since such models do not explicitly estimate the variance, $\sigma^2$, during fitting,  we estimate it using MLE and compute the average log-likelihood as: 
\begin{align*}
    \hat \sigma^2_{MLE} &= \frac{1}{Tn}\sum_{i=1}^n (\y_i - f_{\bm{\theta}}(\x_i))^T (\y_i - f_{\bm{\theta}}(\x_i)), \\
    \frac{1}{n}\log L(\bm{\theta}) &=  - \frac{T}{2} - \frac{T}{2}\log(2\pi \hat\sigma^2_{MLE}).
\end{align*}

For our model, we approximate the average log-likelihood on the test data set based on the MC draws. That is, we estimate the average log-likelihood as: 
\begin{align*}
\z_{ij} &\sim q, \quad\si_{ij} = f_{\bm{\theta}}(\x_i  \lvert \z_{ij}),\\
    \frac{1}{n} \log L(\bm{\theta}) &= \frac{1}{n} \sum_{i=1}^n \log \frac1{k}\sum_{j=1}^k p(\y_i;\hat \si_{ij}). 
\end{align*}

\noindent
\textbf{Calibration.} To further assess the quality of our predictive distribution on the benchmark CGM data set, we propose to evaluate model calibration. We focus on the quality of the estimated marginal distributions for each predicted time point. Given $n$ input sequences, our approach produces approximate predictive distributions based on the MC draws, $\{\hat p(\y_i | \x_i)\}_{i=1}^n$. Based on our copula formulation for multivariate output, we also have direct access to the marginal distributions, $\{[\hat p(y_{ij} | \x_i)]_{j=1}^T\}_{i=1}^n$, each of which is modeled as an IMM. Thus, we can derive approximate marginal cumulative density functions (CDF) as:
$$
    \hat F(y_{ij} | \x_i) = \frac{1}{k} \sum_{i=1}^k F(y_{ij}; \hat \si_{ij}),
$$
where $F$ is the CDF of the chosen univariate base distribution.

To judge the calibration of our model (how close are the estimated confidence intervals to nominal coverage) via a diagnostic plot, we utilize the approach described in \cite{kuleshov2018accurate}. We select a set of $\eta_{jl} \in [0, 1]$ for each $j \in \{1, \dots, T\}$ and a pre-specified mesh $l \in \{1, \dots, L\}$, and calculate
\begin{align*}
    \hat \eta_{jl} = \frac1{n}\sum_{i=1}^n \mathbf{1}\{ \hat F(y_{ij} | \x_i) < \eta_{jl} \}
\end{align*}
Thus, we obtain a set of points $\{(\hat \eta_{jl}, \eta_{jl}) \}_j$ for each $l \in \{1, \dots, L\}$, where $\hat \eta_{jl}$ are empirical frequencies and $\eta_{jl}$ are expected (true) frequencies. Ideally, in the case of perfect calibration, we have that $\hat \eta_{jl} = \eta_{jl}$.

\subsection{Estimation}
\label{sec:transformer}

\textbf{Transformer Architecture.} The Transformer model \cite{vaswani2017attention} consists of a series of attention heads and fully connected layers. To overcome the computational cost of modeling longer prediction intervals, we use a modification of the Transformer architecture that introduces convolutional layers in-between attention blocks and is able to provide multi-output predictions in a single pass \cite{zhouInformerEfficientTransformer2021}. We use the model to predict the parameters of each of the  marginal IMM's, where we consider each stochastic pass through the model to be an MC draw. 

\noindent
\textbf{Feature Embedding.} As outlined in Section~\ref{subsec:data}, the CGM data usually only contains glucose readings together with the timestamps. It is reasonable to assume that each subject adheres to a daily schedule that affects his or her blood glucose levels. Therefore, we utilize time features to enhance our predictions. Also, we add positional and subject encoding.



\begin{figure}[!t]
\centering
\includegraphics[width=140pt]{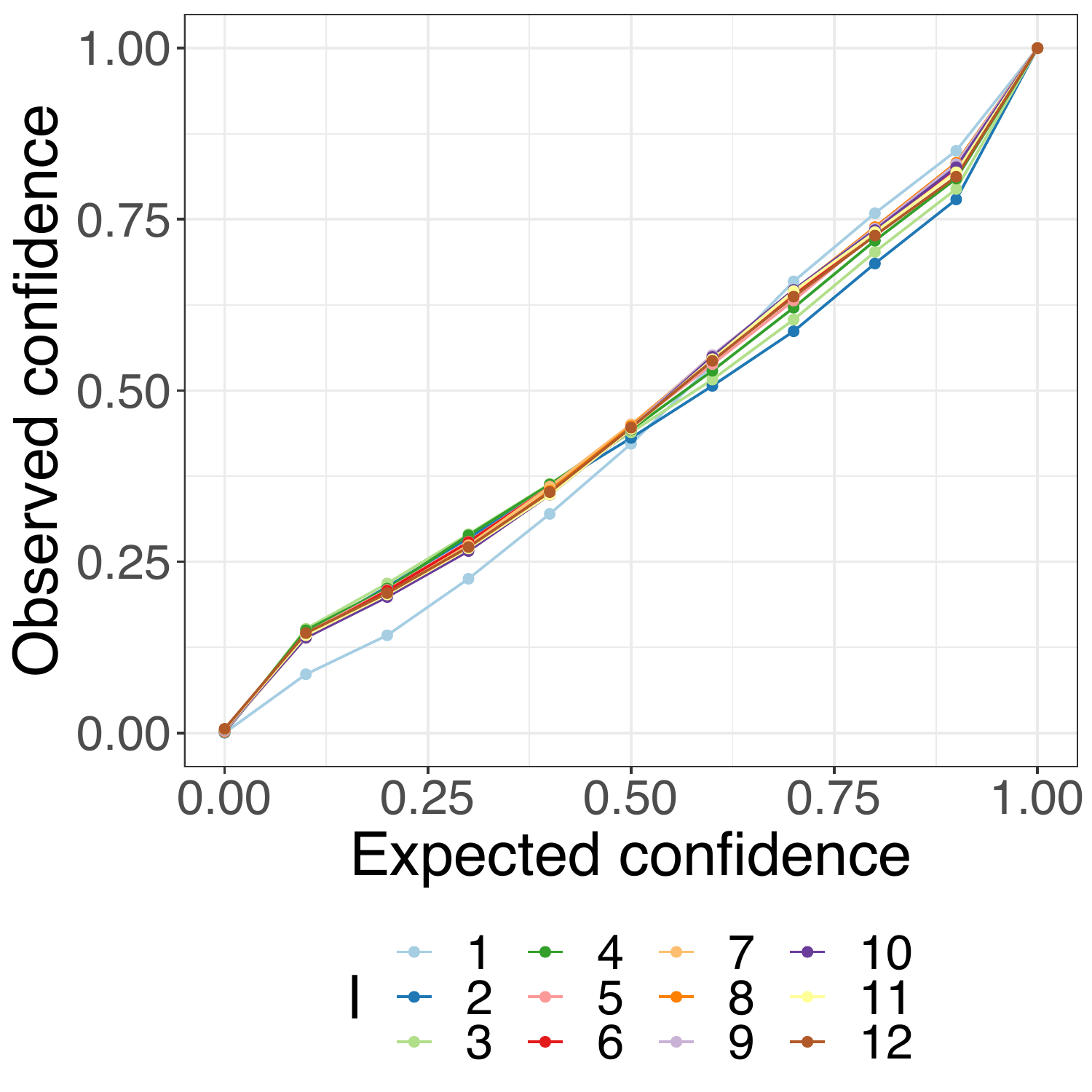}
\caption{Calibration curves depicting $\{(\hat \eta_{jl}, \eta_{jl}) \}_j$ for $l = \{1, \dots, 12\}$ for proposed model on the benchmark CGM data. Perfect calibration would be represented by the $45^\circ$ line.}
\label{fig:calibration}
\end{figure}

\begin{table}[!t]
\centering
\resizebox{110pt}{!}{
\begin{tabular}{c|c|c}
\toprule
Model & \textbf{IMM} & \textbf{Gaussian}  \\
\midrule
APE & 28.63 & 32.68\\
RMSE & 0.1355 & 0.1415\\
Log-likelihood & -0.65 & -4.21 \\
\bottomrule
\end{tabular}}
\caption{APE, RMSE, and likelihood computed on the test set for synthetic data.}
\label{table:prediction1}
\end{table}

\begin{table}[!t]
\centering
\resizebox{250pt}{!}{
\begin{tabular}{c|ccccc}
\toprule
Model & Full & Event & Hypo & Hyper & Likelihood \\
\midrule

ARIMA &  
9.85 / 17.65 & 8.91 / 19.86 & 19.94 / 14.53 & 8.51 / 22.17 & -14.93 \\

RF:Rec &
9.04 / 17.15 & 8.97 / 20.36 & 18.84 / \textbf{12.43} & 8.68 / 23.41 & -14.58 \\

RF:MO & 
10.22 / 18.27 & 8.61 / 19.90 & 21.64 / 17.36 & 7.99 / 21.58 & -15.34 \\

PolySeqMO & 
8.55 / 15.68 & 8.27 / 18.81 & 22.86 / 21.87 & \textbf{6.77 / 18.30} & -15.61 \\

RNN & 
8.17 / 15.67 & 8.29 / 19.37 & 18.72 / 16.26 & 6.99 / 19.22 & -13.50 \\

TFT & 
7.80 / 15.78 & 8.03 / 18.23 & 16.23 / 14.62 & 6.87 / 18.98 & --\\

Our: IMM  & \textbf{7.78} / 15.40  & \textbf{7.89 / 17.85} & \textbf{15.75} / 14.03 & 7.08 / 19.58 & \textbf{-2.67} \\

Our: Gaussian &  
7.82 / \textbf{14.73} & 7.93 / 18.62 & 18.28 / 13.55 & 6.81 / 18.94 & -12.82 \\
\bottomrule
\end{tabular}}
\caption{APE/RMSE for 60-minute prediction window (Full), hypoglycemia, hyperglycemia, and event (hypo-or hyperglycemia), and model log-likelihood on test data.}
\label{table:prediction2}
\end{table}

\section{Performance Evaluation}
\label{sec:results}
\subsection{Data}
\label{subsec:data}
\textbf{Synthetic.} We consider a sample of $2000$ time series for training, $100$ for validation, and $100$ for testing.  We generate the data using a mixture of Gaussian Processes. Each series has the same range of values in the time interval $[-4, 0)$, but in the interval $[0,3]$ takes one of two trajectories: increase or decrease. The resulting distribution has a varying number of modes: one mode on the interval $[-4, 0)$ and two modes on $[0, 3]$. The training curves are depicted in grey in Figure~\ref{fig:prediction}.

\noindent
\textbf{Benchmark CGM.} We use a publicly available CGM data set \cite{foxDeepMultiOutputForecasting2018}, which contains information on glucose levels of $38$ subjects tracked continuously throughout multiple disjoint intervals with the measurement frequency of $5$ minutes. Similar to \cite{armandpourDeepPersonalizedGlucose2021}, we remove periods of drastic fluctuations where the subsequent CGM readings change by more than $40~\text{mg/dL}$. To be compatible with the analysis in \cite{armandpourDeepPersonalizedGlucose2021}, we do not utilize any interpolation techniques for the missing data. The resulting data consists of $399,302$ observations, with an average of 30 (uninterrupted) sequences per subject of length between 200 and 400 observations. We split the data by sequences into the train, validation, and test sets in $20:1:1$~proportion.

\subsection{Model Implementation}
\label{subsec:implement}
We implement our method in PyTorch \cite{paszke2019pytorch}, and use a single NVIDIA RTX 2080 Ti GPU for all experiments. For ablation purposes, we train the model using (i) proposed loss corresponding to the IMM; (ii) mean squared error (MSE) corresponding to the Gaussian assumption. For IMM, we implement the latent variables by introducing the dropout and assuming independence across time. 
The full model implementation is available at \url{https://github.com/mrsergazinov/gluformer}. For the synthetic data, we set the encoder length to 4 and the prediction to 2. For the glucose data, we set the encoder length to 180 (36 hours) and the prediction length to 12 (1 hour).

\subsection{Results}
\label{subsec:prediction}
We measure prediction accuracy using root mean squared error (RMSE) and average percentage error (APE) across prediction intervals. We report the final APE and RMSE as medians of the individual prediction errors across samples. 
We measure the quality of fit as in Section~\ref{subsec:quality}.

\noindent
\textbf{Synthetic.} We compare the proposed IMM to the same architecture trained with the Gaussian assumption. We report our results in Table~\ref{table:prediction1} and plot a sample of predictions in Figure~\ref{fig:prediction}, displaying the predicted densities for the two models. The proposed IMM is more accurate and correctly captures the multi-modality in the predicted densities across time points, providing a better probabilistic fit to the data. 

\noindent
\textbf{Benchmark CGM.} We report both the error on the full data set as well as errors for the periods of hypo- and hyperglycemia \cite{turksoy2013hypoglycemia}. We define a period of hypoglycemia if the glucose level drops below $70~\text{mg/dL}$ in the forecast window and a period of hyperglycemia if the glucose level rises above $180~\text{mg/dL}$. We also measure the model's accuracy during the incidence of either of the events. 
Table~\ref{table:prediction2} summarize the results on the benchmark CGM data for the 60 minutes prediction window. The proposed IMM outperforms the others 3 out of 4 times in APE and 2 out of 4 times in RMSE. In particular, IMM achieves state-of-the-art $7.78$ APE for the most difficult $60$ minute forecast. Table~\ref{table:prediction2} also reports average likelihoods on test data for all methods, except TFT for which it is not available. Our model achieves $6 \times$ better average log-likelihood than the previous methods. We believe this to be indicative of the fact that the true predictive distribution of $\y|\x$ is not similar to Gaussian (likely multi-modal).  We report our calibration plots for each of the $12$ prediction points in the future (the longest prediction length) in Figure~\ref{fig:calibration}. When plotted, a perfectly calibrated model should correspond to a straight line.  We find that the model observed confidence levels are close to the expected levels for all the points, highlighting the model's excellent performance.




\section{Conclusion}
\label{sec:conclusion}
We propose a novel approach to quantify uncertainty in multi-output time series predictions based on the infinite mixture model. We utilize an efficient Transformer architecture for estimation. When applied to public CGM data, our model outperforms existing approaches (ARIMA random forest, RNN, and Transformer) in terms of the achieved log-likelihood on the test data and provides competitive accuracy for various forecast horizons and dangerous event scenarios.

\newpage
\small
\bibliographystyle{IEEEbib}
\bibliography{refs}

\end{document}


\appendix
\section{Transformer Architecture}
\label{app:a}

Let $\mathbf{x_{emb}} \in \mathbb{R}^{t\times d}$ be an embedding of an input $\mathbf{x}$. Then an attention layer, $\mathcal{A}$, is formulated as:
\begin{align*}
    \mathcal{A}(\mathbf{x_{emb}}) &= \text{softmax}\left( \frac{QK^T}{\sqrt{d_k}} \right)V \\
    Q &= W^Q \mathbf{x_{emb}} \\
    K & =W^K \mathbf{x_{emb}}  \\ 
    V &= W^V \mathbf{x_{emb}},
\end{align*}
where the softmax function is applied column-wise. A multi-head attention layer is defined as a combination of the single-head attention layers as:
\begin{align*}
    \mathcal{MA}(\mathbf{x_{emb}}) &= (\mathcal{H}_1(\mathbf{x_{emb}}), \dots, \mathcal{H}_h(\mathbf{x_{emb}}))W^O \\
    \forall i: \mathcal{H}_i(\mathbf{x_{emb}}) &= \mathcal{A}(\mathbf{x_{emb}}) \\
    W^O &\in \mathbb{R}^{hd_v \times d_v},
\end{align*}
where each attention head $\mathcal{H}_i$ has its own weights $W_i^Q, W_i^K$ and $W_i^V$.

We build our Transformer to follow an encoder-decoder architecture. Both encoder and decoder are composed of multi-head attention layers. However, while encoder's attention is focused only on its own input, the decoder has a cross-attention block which incorporates the output of the encoder. Thus, receiving an input $\mathbf{x}$, the encoder part of the model learns a continuous representation $\mathbf{\tilde x}$. Consequently, given $\mathbf{\tilde x}$ and $\mathbf{x}$, the decoder then outputs an estimated mean and log-variance, $(\bm{\hat \mu}, \log \hat \sigma^2)$ of the underlying Gaussian. We introduce the logarithm into the estimation of variance, $\sigma^2$, to prevent the negative variance phenomenon. 

For our task, we find that the main advantage of the Transformer model is $\mathcal{O}(1)$ path between any of the inputs in the sequence. This property allows the Transformer to efficiently transduce signals (measurements) backward and forward in the model, unlike the traditional RNN where the signal has to travel $\mathcal{O}(t)$.

\begin{figure}[t]
\centering
\includegraphics[width=8cm]{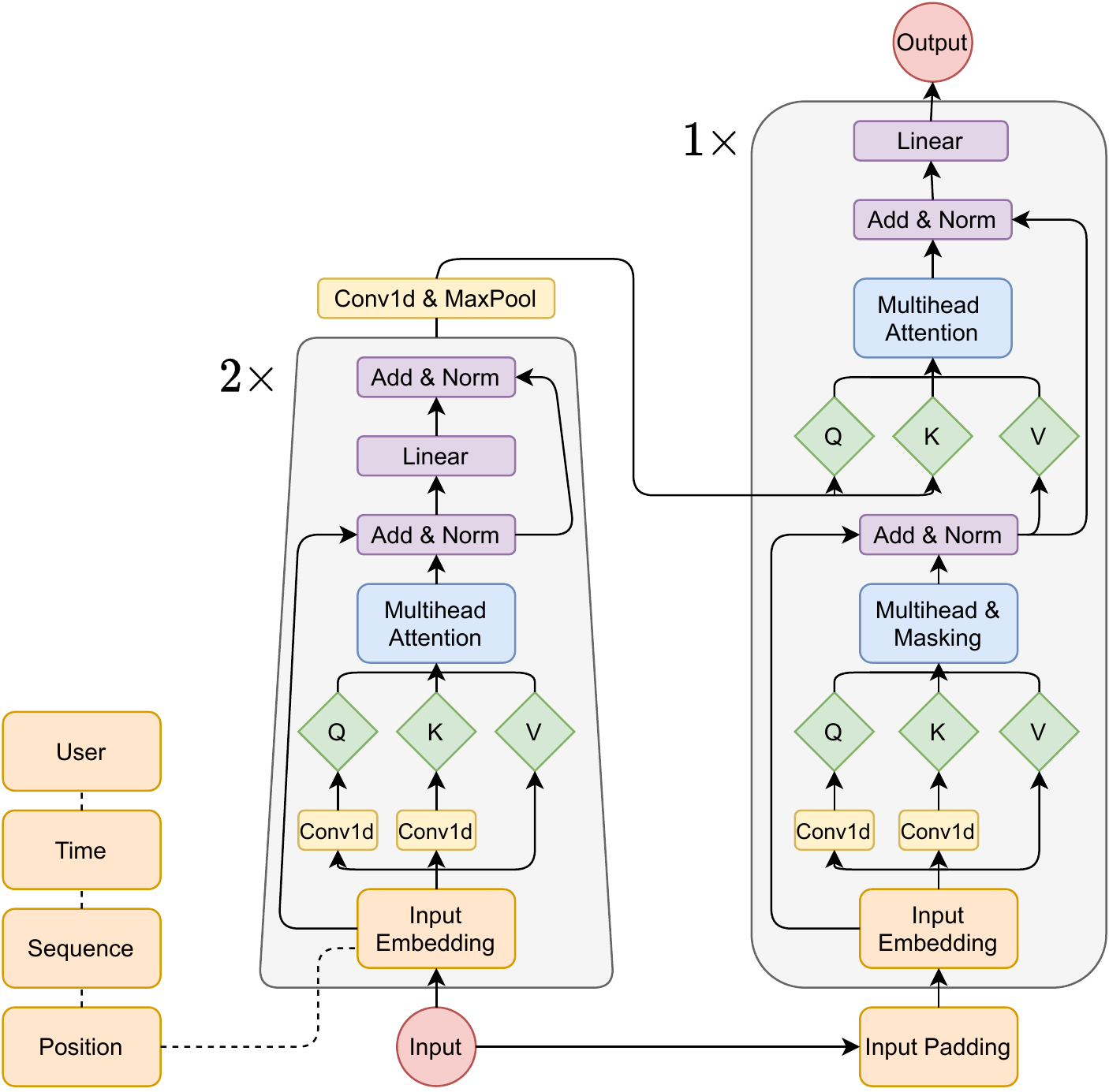}
\caption{Schematic depiction of the proposed architecture based on the modified Transformer model.}
\label{fig:model}
\end{figure}

However, we find that due to the dot-product in the attention, the Transformer has $\mathcal{O}(t^2d_k)$ complexity per layer, where $t$ is the length of the input sequence. This is especially troublesome if several multi-head attention layers are stacked and if longer input is desired, as is the case in our study. To partially circumvent this problem, we implement convolutional layers to reduce the dimensions between the sequential multihead attention layers \cite{li2019enhancing, zhouInformerEfficientTransformer2021}. Let $\mathbf{\tilde x}_j$ be the output of the $j^{th}$ attention block. We propose to reduce it as $\text{MaxPool} (\text{ELU} (\text{Conv1d}(\mathbf{\tilde x}_j)))$ before passing it to the next attention block, where we use $\text{kernel} = 3$ for the Conv1d operation along the time domain and $\text{stride} = 2$ for the MaxPool. We summarize the overall architecture in Figure~\ref{fig:model}.

\section{Feature Embedding}
\label{app:b}
For time features, given an input sequence,~$\mathbf{x}\in\mathbb{R}^t$, and the corresponding time features, $\mathbf{t}\in\mathbb{R}^{t\times d_t}$, we learn separate embedding $\mathbf{x_{emb}} = \mathbf{x} W^E_x $ and $\mathbf{t_{emb}} = \mathbf{t} W^E_t $ and then sum them as $\mathbf{x_{emb}} \gets \mathbf{x_{emb}} + \mathbf{t_{emb}}$, where $W_x^E \in \mathbb{R}^{1 \times d}$ and $W_t^E \in \mathbb{R}^{d_t \times d}$ with $d$ corresponding to the inside model dimensions.

For positional embedding, we utilize the standard approach based on the sine and cosine encoding \cite{vaswani2017attention}. That is we create  $\mathbf{p} \in \mathbb{R}^{t \times d}$ such that 
\begin{align*}
    \mathbf{p}_{i, 2j} &= \sin \frac{i}{10000^{2j/d}} \\
    \mathbf{p}_{i, 2j+1} &= \cos \frac{i}{10000^{2j/d}} \\
    i &\in \{1, \dots, t \} \\
    j &\in \{0, \dots, \lfloor d/2 \rfloor \}
\end{align*}
We then add the positional arguments to the previous embedding as $\mathbf{x_{emb}} \gets \mathbf{p} + \mathbf{x_{emb}}$.

For subject encoding, we concatenate a subject embedding to each of the input sequences. For example, if the sequence $\mathbf{x}$ corresponds to a subject $j$, then we learn a subject embedding $\mathbf{j} \in \mathbb{R}^d$ and concatenate it to the embedding as $\mathbf{x_{emb}} \gets (\mathbf{j}, \mathbf{x_{emb}})$. We summarize our final model schematically in Figure~\ref{fig:model}.

\bibliography{bib.bib}